\documentclass{article} 
\usepackage{iclr2027_conference,times}

\usepackage{amsmath,amsfonts,bm}

\def\eqref#1{equation~\ref{#1}}

\def\1{\bm{1}}

\def\vp{{\bm{p}}}

\def\vr{{\bm{r}}}

\def\vu{{\bm{u}}}
\def\vv{{\bm{v}}}
\def\vw{{\bm{w}}}

\def\vy{{\bm{y}}}

\DeclareMathAlphabet{\mathsfit}{\encodingdefault}{\sfdefault}{m}{sl}
\SetMathAlphabet{\mathsfit}{bold}{\encodingdefault}{\sfdefault}{bx}{n}

\def\gB{{\mathcal{B}}}

\def\gP{{\mathcal{P}}}

\def\gW{{\mathcal{W}}}

\newcommand{\E}{\mathbb{E}}

\newcommand{\R}{\mathbb{R}}

\newcommand{\Var}{\mathrm{Var}}

\DeclareMathOperator*{\argmax}{arg\,max}

\usepackage{amsmath,amssymb,amsthm}
\usepackage{graphicx}
\usepackage{wrapfig}
\usepackage{booktabs}
\usepackage{enumitem}
\usepackage{multirow}
\usepackage{url}
\usepackage{xcolor}
\usepackage{colortbl}
\usepackage{hyperref}
\usepackage{etoc} 
\usepackage{algorithm}
\usepackage{algpseudocode}
\usepackage{tikz}
\usetikzlibrary{arrows.meta,positioning,fit,backgrounds,shapes.geometric,calc}
\usepackage{pifont} 
\usepackage[most]{tcolorbox} 
\newtcolorbox{promptbox}[1]{colback=gray!5!white,colframe=gray!75!black,title=\textbf{#1},breakable,enhanced}

\usepackage{caption}
\definecolor{decisionedge}{HTML}{3D6FA5}
\definecolor{chanceedge}{HTML}{B8860B}
\definecolor{terminaledge}{HTML}{3A6B35}
\definecolor{darkgreen}{RGB}{0,140,0} 

\newtheorem{definition}{Definition}

\newcommand{\Vhat}{\hat{V}}
\newcommand{\Qhat}{\hat{Q}}
\newcommand{\CCS}{\mathrm{CCS}}
\newcommand{\prune}{\mathrm{prune}}
\newcommand{\msum}{\oplus}
\newcommand{\fth}{f_\theta}

\title{\method: Multi-Objective Agentic Workflow Generation}

\author{Yining Lu\thanks{Work done during an internship at IBM} \\
University of Notre Dame
\And
Aurelie Lozano \\
IBM
\And
Xi Yang \\
IBM
\And
Naoki Abe \\
IBM
\AND
Yu Deng \\
IBM
\And
Meng Jiang \\
University of Notre Dame
}

\newcommand{\method}{\text{MoFlow}}
\newcommand{\pmstd}[1]{{\scriptsize\,$\pm$#1}}

\newif\ifshowcc
\showcctrue                     
\definecolor{cccolor}{HTML}{B02418}
\newcounter{cccount}

\newif\ifarxiv
\arxivtrue
\ifarxiv \iclrfinalcopy \fi

\begin{document}

\maketitle
\ifarxiv \lhead{\method: Multi-Objective Agentic Workflow Generation} \fi
\etocdepthtag.toc{mainmatter}

\begin{abstract}
We study the generation of agentic workflows that jointly optimize multiple objectives, such as accuracy, cost, latency, robustness, and consistency. Existing methods for workflow generation typically optimize accuracy alone or a weighted sum of objectives, so each trained generator commits to one fixed trade-off and must be retrained from scratch when preferences change.
To alleviate this, we propose \method{}, which generates workflows optimized across varied preferences. Specifically, \method{} formulates workflow generation as a multi-objective Markov decision process and solves it by leveraging Convex-Hull Monte Carlo Tree Search with optimistic set-valued backups, where every node stores a set of reachable trade-offs rather than one weighted score. A single search thus approximately covers the Pareto front, from which \method{} can return a workflow for any preference by lookup without retraining. We evaluate \method{} against six strong baselines on six benchmarks spanning mathematics, code, and question answering. Since the baselines are single-scalar optimizers by design, an apples-to-apples comparison is difficult. We instead adopt an evaluation setup that favors the baselines, in that they are rerun for each testing preference, which \method{} never sees. Even under this stringent setup, \method{} achieves the highest average hypervolume.\footnote{Hypervolume measures how well a set of solutions covers the achievable trade-offs among objectives. We open source our code at \href{https://github.com/yining610/MoFlow}{https://github.com/yining610/MoFlow}.}
\end{abstract}

\section{Introduction}

An agentic workflow, a structured graph that composes LLM calls and task-specific operations, can substantially outperform a single prompt on complex, reasoning-heavy tasks \citep{yue-etal-2025-masrouter, li2026assemble, nielsen2026learning}. A growing body of work therefore automates workflow construction by searching over candidate graphs and optimizing them for a single performance metric such as accuracy, as in AFlow \citep{zhang2025aflow} and its successors \citep{wang2025scoreflowmasteringllmagent,zhao2025a2flowautomatingagenticworkflow, chen2026lemonlearningexecutablemultiagent, zhang2026flowsteeragentsdesigningagentic, hu2026evomasevolutionarygenerationmultiagent}. Real-world deployment, however, is inherently multi-objective: beyond accuracy alone, workflows often need to be cheap, fast, robust to noisy input, and consistent across repeated runs \citep{rabanser2026towards}.

These objectives often compete, and different real-world deployments may prioritize them differently. For instance, an enterprise workflow prefers low cost and latency, a scientific one robustness and consistency. The right weighting is rarely known in advance, so a practical generator needs to discover a diverse set of trade-offs from which users can choose according to their preferences. Yet no existing generator directly supports this goal. Accuracy-driven generators return a workflow optimized for a single objective \citep{zhang2025aflow, hu2025automated, 11420713, zhang2026flowsteeragentsdesigningagentic,zhang2026skillflowflowdrivenrecursiveskill}, while the few that look beyond accuracy collapse a handful of objectives, typically accuracy and cost, into a single scalar by simple weighted scalarization \citep{kong2026workflowr1groupsubsequencepolicy,zhang2025multiagent, xu2025robustflowrobustagenticworkflow, hu2026evomasevolutionarygenerationmultiagent}. Either way the generator commits to one point on the Pareto front, and reaching a different point requires retraining the generator from scratch. We therefore ask: \emph{can a workflow generator span the Pareto front of workflows in a single run?}

Our answer is yes: rather than settle on one trade-off, a search can remember every trade-off worth making. Representative generators such as AFlow \citep{zhang2025aflow} run Monte Carlo Tree Search (MCTS) \citep{kocsis2006bandit, browne2012survey, silver2016mastering, schrittwieser2020mastering} with a scalar per node, collapsing all trade-offs into one number, and generate each workflow as a whole, making individual design choices hard to trace. We therefore propose \textbf{\method{}}, which keeps the MCTS search but replaces each node's scalar with a convex coverage set (CCS) \citep{ICAPS20paper100}, the best trade-offs attainable under each preference, and grows each workflow through stepwise graph edits rather than whole-workflow revision. Formally, we formulate workflow generation as a multi-objective Markov decision process (MOMDP) \citep{10.1007/11672142_26}, under which Convex-Hull Value Iteration (CHVI) \citep{10.1145/1390156.1390162} propagates the per-node CCSs to the root, so a single search approximates a Pareto front across preferences.

\begin{figure}[t]
  \centering
  \includegraphics[width=\textwidth]{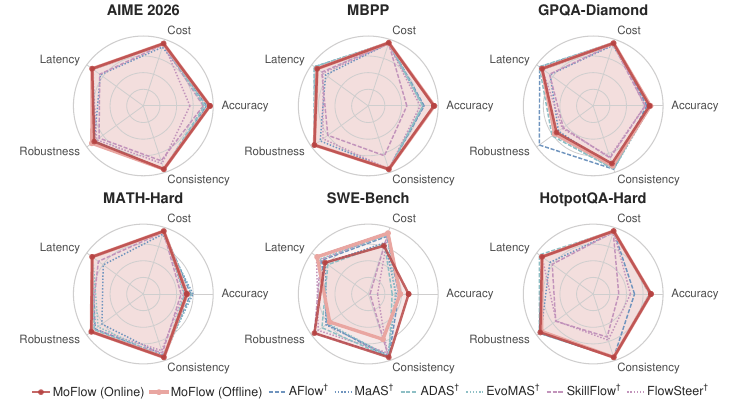}
  \caption{\textbf{Per-objective coverage on the six benchmarks: across eleven testing preferences, \method{} matches or beats every baseline on all five objectives on most tasks.} Each axis shows the best value a method attains across its eleven generated workflows, with cost and latency inverted so that outward is always better. Each baseline$^{\dagger}$ is rerun once per testing preference, whereas \method{} serves all eleven preferences from one search that never saw them.}
  \label{fig:by-dataset}
\end{figure}

Replacing scalars with sets dictates what the method searches for and stores, but not how it allocates its effort. One challenge is where in preference space to search. Each \method{} trial samples a user preference as a context and, through Contextual Zooming for Trees (CZT) \citep{JMLR:v15:slivkins14a}, conditions action selection on it at every node, so a single tree learns how the best action shifts with the preference. The other challenge is that a partial workflow cannot be run, yet the search must evaluate it to decide where to explore next. Motivated by evidence that workflow performance can be predicted without execution \citep{zhang2025gnnspredictorsagenticworkflow}, \method{} trains a graph neural network (GNN) on executed workflows to evaluate partial ones directly. How the GNN is fit yields the two variants we report throughout. Online \method{} fits the GNN during the search on workflows it executes, requiring no prior data (more general but more expensive). Offline \method{} pretrains the GNN on executions from previous online runs and searches with no execution at all (cheaper but less general).

Compared with six strong baselines across six benchmarks spanning mathematics, code, and question answering, on five objectives (we want a workflow to be accurate, consistent, robust, yet cheap and fast), the two \method{} variants attain the best hypervolume on four benchmarks and trail the best by under 1\% on the other two. Averaged over the six benchmarks, both \method{} variants rank first and second, ahead of every baseline, improving the best baseline's hypervolume by 10.9\% (online) and 12.8\% (offline, at least 17 times cheaper). The comparison deliberately favors the baselines. None was built for more than two objectives, so we adapt each to optimize the scalarized reward over all five, rerunning it once per testing preference so that every baseline learns the exact weights it is tested on.\footnote{Hereafter, $^{\dagger}$ denotes an adapted baseline, which sees the testing preference weights during training.} \method{}, by contrast, sees none of the testing preferences. Even so, it matches or beats every baseline on each of the five objectives across most benchmarks, as \autoref{fig:by-dataset} previews.

In summary, we (i) formulate workflow generation as a stepwise MOMDP where an LLM proposes and realizes each graph edit (\S\ref{sec:background}); (ii) solve it with a multi-objective tree search whose nodes store CCSs, so a single search approximates a Pareto front that serves any preference (\S\ref{ssec:nodes}); (iii) make the search practical by conditioning action selection on the sampled preference via CZT and valuing partial workflows with an uncertainty-gated GNN (\S\ref{ssec:czt}); and (iv) validate \method{} against six strong baselines across six benchmarks, five objectives, and four base models (\S\ref{sec:experiments}).

\section{Related Work}
\label{sec:related}

\begin{table}[ht]
  \centering
  \small
  \setlength{\tabcolsep}{2.5pt}
  \caption{\textbf{Comparison of \method{} with the baselines.}
  $2{\to}1$: two objectives scalarized into one. $D$: number of objectives ($D{=}5$ in our experiments). \emph{Pareto}: one run returns the Pareto front. \emph{Stepwise}: the search builds workflows step by step.
  $E$ is the cost of one full workflow evaluation (many LLM calls) and $\ell$ that of a single call ($\ell\ll E$), over $N$ search iterations, $N_{\text{tr}}$ training episodes, or horizon $H$. $^{\ddagger}$Plus LLM-policy gradient updates. \emph{Inference}: cost to serve a new preference (a weight vector over the objectives) by rerunning the search or training. \method{} instead answers any preference by lookup. EvoMAS has no training phase. Its search runs at test time as queries arrive, so a new preference reruns it in full.
  \ding{51}~yes, \ding{55}~no, \ding{119}~partial.}
  \label{table:compare}
  \begin{tabular}{@{}llcccll@{}}
  \toprule
  Method & Family & Objectives & Pareto & Stepwise & Training & Inference \\
  \midrule
  AFlow \citep{zhang2025aflow}          & \multirow{2}{*}{Monolithic} & 1            & \ding{55} & \ding{55}  & $O(NE)$ & $O(NE)$ re-search \\
  MaAS \citep{zhang2025multiagent}      &                                 & 2$\to$1 & \ding{55} & \ding{55} & $O(N_{\text{tr}}E)$ & $O(N_{\text{tr}}E)$ retrain \\
  \midrule
  FlowSteer \citep{zhang2026flowsteeragentsdesigningagentic} & \multirow{2}{*}{Constructive} & 1 & \ding{55} & \ding{51} & $O(N_{\text{tr}}H\ell)^{\ddagger}$ & $O(N_{\text{tr}}H\ell)$ retrain \\
  SkillFlow \citep{zhang2026skillflowflowdrivenrecursiveskill} &                          & 1 & \ding{55} & \ding{119} & $O(N_{\text{tr}}H\ell)^{\ddagger}$ & $O(N_{\text{tr}}H\ell)$ retrain \\
  \midrule
  EvoMAS \citep{hu2026evomasevolutionarygenerationmultiagent} & \multirow{2}{*}{Evolutionary} & 2$\to$1 & \ding{55} & \ding{55}  & \multicolumn{1}{l}{\;\;\;\;\;\;--} & $O(NE)$ re-search \\
  ADAS \citep{hu2025automated}          &                               & 1            & \ding{55} & \ding{55}  & $O(NE)$ & $O(NE)$ re-search \\
  \midrule
  \textbf{\method{}} (Online)  & \multirow{2}{*}{Constructive} & $D$ & \ding{51} & \ding{51} & $O(NE)$ & \multirow{2}{*}{lookup (no LLM)} \\
  \textbf{\method{}} (Offline) & & $D$ & \ding{51} & \ding{51} & $O(NH\ell)$ & \\
  \bottomrule
  \end{tabular}
  \end{table}
\subsection{Automatic Workflow Generation}

We organize automatic workflow generators into three families (\autoref{table:compare}). \method{} does not sit squarely in any of these three categories: it is polylithic, not monolithic (though it is MCTS-based as in AFlow), constructive but multi-objective (unlike other constructive methods), and learning-based rather than evolutionary.

\paragraph{Monolithic workflow generation.} A first line of work treats the complete workflow as the unit of generation and revision, scoring each candidate by executing it end to end. The most representative is AFlow \citep{zhang2025aflow}, which also applies MCTS: each tree node holds a full workflow, and each expansion invokes an LLM optimizer to revise that workflow as a whole. Executing the revised workflow then gives the node a single accuracy score. The search thus returns one workflow rather than a Pareto front, and credit goes to the entire revision, never to the individual edits within it. Both limitations extend across the family, whether a method trains against a scalar reward \citep{fan2024workflowllmenhancingworkfloworchestration, wang2025scoreflowmasteringllmagent, gao2025flowreasonerreinforcingquerylevelmetaagents, xu-etal-2026-comfyui}, samples from a learned architecture distribution as MaAS does \citep{zhang2025multiagent}, or composes reusable operators \citep{zhao2025a2flowautomatingagenticworkflow}.

\paragraph{Constructive workflow generation.} A more recent line addresses credit assignment by constructing workflows through stepwise decisions. FlowSteer \citep{zhang2026flowsteeragentsdesigningagentic} builds a workflow through atomic canvas edits, SkillFlow \citep{zhang2026skillflowflowdrivenrecursiveskill} through stepwise invocations of recursively evolving skills, and related methods likewise assign credit or capabilities to individual steps \citep{zhuge2024gptswarm, niu2025flow, chen2026lemonlearningexecutablemultiagent,kong2026workflowr1groupsubsequencepolicy,wang2026learningcomposecrossdomainagentic,yuan-etal-2026-bayesflow}. Although this finer granularity improves interpretability, it does not by itself preserve trade-offs among objectives. A few methods do consider a second objective beyond accuracy, either token cost \citep{wang-etal-2025-agentdropout, zhang2025cut} or robustness \citep{xu2025robustflowrobustagenticworkflow}, yet all of them collapse the two into a single scalar. In contrast, we keep the stepwise search but propagate each node's CCS with CHVI backups, so each edit receives separate credit along every objective.

\paragraph{Evolutionary workflow generation.} A third line grows workflows by iterative variation and selection rather than by planning \citep{zhang2025evoflowevolvingdiverseagentic, wang-etal-2025-evoagentx, liu2026sewselfevolvingagenticworkflows, Xiao2026AgentEvoCA}. EvoMAS \citep{hu2026evomasevolutionarygenerationmultiagent} mutates and recombines configurations, while ADAS \citep{hu2025automated} has a meta agent write new designs in code, building on a growing archive of prior ones. These methods explore a rich space of complete workflows, but a single scalar still decides what survives each round, so the search again converges to one region of the front.

Two additional lines of work underpin \method{} directly. \citet{rabanser2026towards} propose four dimensions for the scientific evaluation of AI agents. We adopt two of them (robustness and consistency) alongside accuracy, cost, and latency to form our five objectives. \citet{zhang2025gnnspredictorsagenticworkflow} argue that workflow performance is predictable from the workflow graph and its prompts alone, without execution, an idea extended by \citet{guan2025glowgraphlanguagecoreasoningagentic} and \citet{feng2026rewardflow}. Because partial workflows cannot yet be executed, we use a GNN to evaluate them directly from their graph representation.

\subsection{Tree Search for Multi-Objective Reinforcement Learning}
\label{sec:related_tree_search_morl}

Solving an MOMDP means recovering a Pareto front of policies \citep{Roijers_2013, NEURIPS2019_4a46fbfc, 3545946.3598872, basaklar2023pdmorl, zhang2023hypervolume, liu2025paretosetlearningmultiobjective, li2025how, liu2025efficient}, and multi-objective MCTS (MOMCTS) pursues this goal by adaptively allocating simulations across the front. The original MOMCTS \citep{pmlr-v25-wang12b} keeps a single global front at the root, which cannot express purely local trade-offs \citep{ICAPS20paper100, hayes2022montecarlotreesearch}. Later methods instead maintain a local Pareto set at every node, following Pareto Q-learning \citep{JMLR:v15:vanmoffaert14a}, and differ in action selection: Pareto dominance \citep{Chen_2019, 6707036}, hypervolume-based UCB \citep{auer2002finite, 6633621, 6872573, 7743851}, or Chebyshev scalarization \citep{8107102}. All of these assume deterministic transitions, where the chosen action determines the successor, whereas ours are stochastic because an LLM can instantiate the same action into different workflows. We build on Convex-Hull MCTS (CHMCTS) \citep{ICAPS20paper100}, which already handles stochastic transitions via CHVI within Trial-based Heuristic Tree Search \citep{keller2013trial}, and extend it with an LLM that proposes and realizes actions, so neither the action set nor the transition model is fixed.

\section{\method{}: Multi-Objective Agentic Workflow Generation}
\label{sec:algorithm}\label{ssec:loop}

\subsection{Workflow Construction as a Multi-Objective MDP}
\label{sec:background}
Building a workflow requires a sequence of actions whose outcomes may trade off against one another, so we formalize it as a finite-horizon multi-objective Markov decision process (MOMDP) $M=\langle S,A,R,T,\bar{s},H\rangle$. An MOMDP replaces the scalar reward of an ordinary MDP with a vector of $D$ objectives, $R:S\times A\times S\to\R^D$. In our setting, this reward is obtained only at the terminal state by executing the complete workflow, and we cast every objective as maximization, so cost and latency are negated in $R$. The remaining elements are standard: $S$ is the state space of partial workflows, $A$ the action space of atomic edits, $T$ the transition kernel, $\bar{s}$ the empty workflow, and $H$ the horizon. Because rewards are vectors, a policy $\pi$ is evaluated by a vector value $\vv^\pi(s)=\E\big[\sum_{t}R(s_t,a_t,s_{t+1})\mid s_0=s,\pi\big]\in\R^D$, and a user \emph{preference} is a weight vector $\vw\in\Delta^{D-1}$ that scalarizes it as $\vw^\top\vv$. Two value vectors need not be comparable (one workflow may be more accurate while another is cheaper), so solving an MOMDP means ranking outcomes by \emph{Pareto dominance} and recovering the entire \emph{Pareto front} of non-dominated trade-offs in one run.

\begin{definition}[Pareto dominance and Pareto front]\label{def:pareto}
A vector $\vu$ \emph{dominates} $\vv$, $\vu\succ\vv$, if $u_i\ge v_i$ for every objective $i$ and $u_j>v_j$ for at least one $j$. Given a set $Y\subset\R^D$ of achievable value vectors, its \emph{Pareto front} is the non-dominated subset $P=\{\,\vr\in Y:\nexists\, \vr'\in Y \text{ with } \vr'\succ \vr\,\}$.\end{definition}

We measure the quality of a front $P$ by its \emph{hypervolume}, the volume of objective space that $P$ dominates above a fixed reference point $\vr_0$,
\[
\mathrm{HV}(P;\vr_0)=\Lambda\big(\{\,\vy\in\R^D:\exists\,\vp\in P,\ \vr_0\preceq\vy\preceq\vp\,\}\big),
\]
with $\Lambda$ the Lebesgue measure and larger better. Since we only consider linear preferences, which scalarize the value vector as $\vw^\top\vv$, it suffices to keep a \emph{convex coverage set} (CCS), a subset $\CCS\subseteq P$ that still attains the best scalarized value under every preference, $\max_{\vv\in\CCS}\vw^\top\vv=\max_{\vv\in Y}\vw^\top\vv$ for all $\vw\in\Delta^{D-1}$ \citep{Roijers_2013}. Our backups prune only Pareto-dominated vectors, so each node in fact stores the full front, a superset of the minimal CCS (Appendix~\ref{app:algo}).

\begin{figure}[t]
  \centering
  \includegraphics[width=\textwidth]{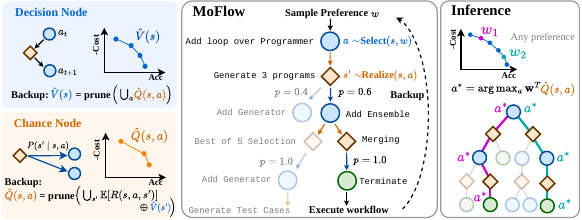}
  \caption{\textbf{Overall \method{} framework.} \method{} searches once, then answers any preference by lookup. Each trial samples a preference $\vw$ and walks the tree as follows. At a \textcolor{decisionedge}{\textbf{decision node}}, the trial selects an action for $\vw$; at a \textcolor{chanceedge}{\textbf{chance node}}, an LLM realizes that action into at most $N_r$ distinct outcomes, each visited with empirical probability $p$ estimating $T(s^\prime\mid s,a)$. Decision and chance nodes alternate until a \textcolor{terminaledge}{\textbf{terminal}} action completes the workflow; \method{} then executes it and backs the result up the visited path (faded nodes are branches the current trial does not visit). Every node therefore stores a convex coverage set ($\Vhat(s)$ at decision nodes, $\Qhat(s,a)$ at chance nodes), so a single tree can unroll different workflows for different preferences such as $\vw_1$ and $\vw_2$ (\emph{Inference} panel).}
  \label{fig:tree}
\end{figure}

\begin{algorithm}[t]
  \small
  \caption{Algorithm of \method{} for a single trial.}\label{alg:trial}
  \begin{algorithmic}[1]
  \State $\vw \sim \Delta^{D-1}$ \Comment{\textcolor{darkgreen}{the CZT context, resampled at each trial}}
  \State $s \gets \bar{s}$
  \While{$s$ is not terminal \textbf{and} $s$ is already in the tree} \Comment{\textcolor{darkgreen}{a trial ends at a terminal or newly reached state}}
    \State $\widehat A \gets \textsc{Propose}(s)$;\quad $\gB \gets$ the relevant balls of $\widehat A$ at $\vw$ \Comment{\textcolor{darkgreen}{at the decision node $s$}}
    \State $a \gets \textsc{Select}(s,\vw)$, the action of $\arg\max_{B\in\gB} I(B)$ \Comment{\textcolor{darkgreen}{Eq.~\ref{eq:czt-index}}}
    \State $s^\prime \sim \textsc{Realize}(s, a)$;\quad update frequency $p$;\quad $s \gets s^\prime$ \Comment{\textcolor{darkgreen}{at the chance node $(s,a)$}}
  \EndWhile
  \State $\Vhat_{\text{leaf}} \gets$ execute $s$ if terminal, else $\fth(s)$;\quad $u \gets \max_{\vv\in\Vhat_{\text{leaf}}}\vw^\top\vv$
  \For{$(s,a,B,s^\prime)$ along the visited path, back to the root}
    \State $\Qhat(s,a)\gets$ Eq.~\ref{eq:chance-backup};\quad $\Vhat(s)\gets$ Eq.~\ref{eq:decision-backup} \Comment{\textcolor{darkgreen}{CHVI backup}}
    \State fold $u$ into $\nu(B)$;\quad split $B$ at $\vw$ if $\mathrm{conf}(B)\le r(B)$ \Comment{\textcolor{darkgreen}{the contextual zooming step}}
  \EndFor
  \end{algorithmic}
\end{algorithm} 

\subsection{\method{} Search and Inference}
\label{ssec:nodes}

\method{} solves the MOMDP by MCTS search in which each node is valued by the set of trade-offs reachable from it rather than by a single scalar. Because each trial samples a fresh preference $\vw$ to guide action selection, a single search approximately covers the front. Given any preference $\vw$ (even one never sampled during search), the resulting tree serves the preference by lookup without retraining. Algorithm~\ref{alg:trial} gives an overview of the single trial procedure.

The search is driven by two LLM roles. A \emph{proposer} suggests at most $b$ candidate edits at each partial workflow, and a \emph{realizer} instantiates the selected edit into at most $N_r$ distinct outcomes. The tree accordingly alternates two kinds of node: a \emph{decision node}, where the search chooses among the proposer's suggested actions, and a \emph{chance node}, where it observes which successor the realizer produces. Both store a CCS of value vectors, preserving the trade-offs among objectives rather than collapsing them to a scalar. \autoref{fig:tree} illustrates these sets with two objectives, accuracy and cost.

\begin{definition}[Decision node]
A \emph{decision node} is a partial workflow $s\in S$ that stores a CCS estimate $\Vhat(s)$. On each visit, the proposer returns at most $b$ candidate actions drawn from an operator pool, and the search selects one for the current preference (\S\ref{ssec:czt}). A child chance node is created for the selected action. The root is the empty workflow $\bar{s}$.
\end{definition}

\begin{definition}[Chance node]
A \emph{chance node} is a state-action pair $(s,a)\in S\times A$ that stores a CCS estimate $\Qhat(s,a)$. Because the realizer may produce a different successor on each visit, a single chance node can branch into multiple child decision nodes $s^\prime$, and each branch's empirical visit frequency $p$ estimates $T(s^\prime\mid s,a)$ for sampling at inference.
\end{definition}

Separating the chosen action from its realization turns stochasticity into counted branches rather than score noise. CHVI maintains both CCS estimates by backing up the entire front after each trial. Its exact chance backup is the probability-weighted Minkowski sum over successors, which we replace with a cheaper \emph{optimistic} union that keeps every vector achievable under at least one realization:
\begin{equation}
\Qhat(s,a) = \prune\Big(\bigcup_{s^\prime} \big(\E[R(s,a,s^\prime)]\msum \Vhat(s^\prime)\big)\Big),
\label{eq:chance-backup}
\end{equation}
where $\prune$ keeps only the non-dominated vectors. A workflow can only be run once it is complete, so $\E[R]$ is estimated only at terminal states, from repeated executions of the completed workflow. A partial workflow carries no reward and contributes only $\Vhat(s^\prime)$, estimated by a GNN value model introduced below. For a decision node, CHVI unions and prunes its actions' CCS estimates:
\begin{equation}
\Vhat(s) = \prune \Big(\bigcup_a \Qhat(s,a)\Big).
\label{eq:decision-backup}
\end{equation}
Once a trial reaches its leaf and the completed workflow is evaluated, \method{} applies the two backups in alternation back up the visited path to the root. Each node's CCS then covers the best trade-offs its completions can achieve, so a single trial improves the answer for every preference, not just the one it sampled. Appendix~\ref{app:algo} explains why the chance backup uses an optimistic union rather than the exact probability-weighted sum.

When the leaf is still partial, some objectives require execution to evaluate. A GNN value model $\fth$ predicts these objectives along with its own uncertainty. If the uncertainty exceeds a gate, \method{} executes a few cheap completions and replaces the prediction with the non-dominated returns (details in Appendix~\ref{app:setup-details}). These returns also fill a replay buffer that online \method{} uses to update $\fth$ as the search proceeds. Offline \method{} instead pretrains $\fth$ on executions logged from online runs, then searches without any execution.

At inference, given a preference $\vw$, \method{} walks the built tree from the root, selecting the $\vw$-optimal action $a^\ast=\argmax_{a}\max_{\vv\in\Qhat(s,a)}\vw^\top\vv$ at each decision node and sampling a successor by its empirical visit frequency $p$ at each chance node (\emph{Inference} panel of \autoref{fig:tree}). The entire walk reads stored CCS estimates, so serving a new preference requires no additional LLM calls.

\begin{wrapfigure}[17]{r}{0.43\textwidth}
  \vspace{-19pt}
  \centering
  \includegraphics[width=0.34\textwidth]{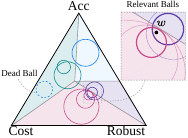}
  \caption{\textbf{Contextual zooming at one decision node} over the preference simplex for $D{=}3$ objectives. Each ball ties one action (its color) to a preference region. The background marks the true optimal action the search never sees. Finer balls tell nearby preferences apart, so refinement drifts toward region boundaries while interiors stay coarse.}
  \label{fig:czt}
\end{wrapfigure}
\subsection{Contextual Zooming for Trees}
\label{ssec:czt}

Preference enters the search only at action selection. Each decision node is a contextual bandit \citep{NIPS2011_e1d5be1c} whose arms are the proposed actions and whose winning arm changes across the simplex. A single ranking of the arms for all $\vw$ incurs regret $\Omega(N)$ over $N$ trials \citep{ICAPS20paper100}, and a uniform grid over $\Delta^{D-1}$ achieves resolution everywhere at a cost exponential in $D$. Contextual Zooming for Trees (CZT) \citep{ICAPS20paper100} instead stays coarse where one action dominates and refines only near the boundaries where the winner changes. Its regret scales with the zooming dimension $c$ of those boundaries rather than the ambient dimension $D-1$.

Concretely, CZT covers the preference simplex with balls, each tying one action to a preference region (\autoref{fig:czt}). Only the finest ball of each legal action covering the sampled $\vw$ is \emph{relevant}, and these compete by the optimistic index:
\begin{align}
&I_{\mathrm{pre}}(B)=\underbrace{\nu(B)}_{\text{reward}}+\underbrace{r(B)}_{\text{radius}}+\underbrace{\mathrm{conf}(B)}_{\text{confidence}},\nonumber\\
&I(B)=r(B)+\min_{B'\ \mathrm{active}}\big[I_{\mathrm{pre}}(B')+d_s(B,B')\big].
\label{eq:czt-index}
\end{align}
An action is more likely to be selected when its relevant ball $B$ has a high reward $\nu(B)$ (the average scalarized reward under the sampled $\vw$), few visits $n(B)$ (a wide $\mathrm{conf}(B)=\kappa\sqrt{\log N/(1+n(B))}$, with $N$ the total trial budget), or a coarse radius $r(B)$. Once a ball has been selected enough times to be confident at its own scale (i.e., $\mathrm{conf}(B)\le r(B)$), we split it into a half-radius child centered at the current $\vw$. A ball whose action keeps losing at the preferences it covers eventually stops being selected and goes \emph{dead}. Unlike classical CHMCTS, our proposer redraws the action set at every visit. Because the operator pool is fixed, the same operators reappear across visits but with different free-text roles, adding diversity without creating blind spots (Appendix~\ref{app:method}).

\begin{figure}[t]
  \centering
  \includegraphics[width=\textwidth]{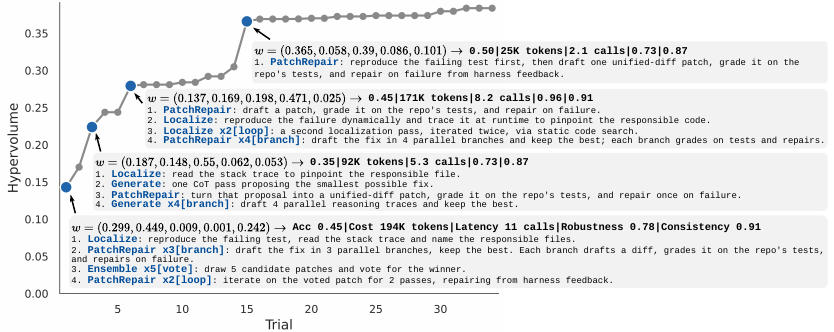}
  \caption{\textbf{One \method{} search on SWE-Bench.} The y-axis shows the in-sample hypervolume of non-dominated workflows executed up to each trial. Each labeled example is a workflow generated from its sampled $\vw$, with objectives listed as accuracy $|$ cost $|$ latency $|$ robustness $|$ consistency.}
  \label{fig:example}
\end{figure}

\section{Experimental Setup}
\label{ssec:setup}

\paragraph{Benchmarks.}
We evaluate on six benchmarks spanning three task families for which agentic workflows are typically built. For mathematics we use AIME 2026 \citep{dekoninck2026benchmarksmatharenaevaluationplatform} and MATH-Hard, the Level-5 subset of MATH \citep{hendrycks2021measuring}. For code we use the sanitized subset of MBPP \citep{austin2021programsynthesislargelanguage} and SWE-Bench Lite \citep{jimenez2024swebench}. For question answering we use GPQA-Diamond \citep{rein2024gpqa} and HotpotQA-Hard \citep{yang-etal-2018-hotpotqa}. Each benchmark is split into a build set used during search and a disjoint held-out set on which we report all results (10 build\,/\,20 held-out for AIME 2026, $20/50$ elsewhere).\footnote{Each workflow runs six times per problem, so evaluating \method{} on a 50-problem held-out set takes up to 3{,}300 executions (over 10{,}000 LLM calls on SWE-Bench alone), already matching AFlow's total evaluation budget. The 10--20 problem build sets still suffice for generalization, as shown in \autoref{table:cost} (Appendix~\ref{ssec:modes-exp}).}

\paragraph{Objectives and metrics.}
Each workflow is evaluated on five objectives: accuracy, cost, latency, robustness, and consistency. We execute every problem six times, with $J=3$ repeated runs of the original problem and one run of each of its $m=3$ paraphrases. We report accuracy (task success), cost (total tokens per problem), and latency (sequential LLM calls) by averaging over the repeated runs. Robustness and consistency share the formula $\mathrm{clip}(1-4\Var(\mathrm{acc}),0,1)$ and differ only in where the accuracy variance comes from. Consistency measures it across the repeated runs, while robustness measures it across the paraphrases. A good workflow attains high accuracy, robustness, and consistency at low cost and latency.\footnote{The five objectives capture distinct properties. Even the closest pair, robustness and consistency, correlates at only 0.63 over 132 served workflows, from $-0.13$ on GPQA to 0.87 on SWE-Bench. We measure latency as sequential LLM calls rather than wall-clock time because it varies with provider, hardware, and load.} Each method generates one workflow per testing preference, and we evaluate it by the hypervolume of these workflows on the held-out set with respect to the nadir point $\vr_0=[0,-3{\times}10^5,-20,0,0]$.

\paragraph{Baselines.}
We compare against six recent workflow generators, two from each family in \S\ref{sec:related}: AFlow \citep{zhang2025aflow}, MaAS \citep{zhang2025multiagent}, FlowSteer \citep{zhang2026flowsteeragentsdesigningagentic}, SkillFlow \citep{zhang2026skillflowflowdrivenrecursiveskill}, ADAS \citep{hu2025automated}, and EvoMAS \citep{hu2026evomasevolutionarygenerationmultiagent}. All methods are evaluated on the same eleven testing preferences $\vw$, spanning the simplex from corners to center: the five permutations of $[1,0,0,0,0]$, the five permutations of $[0.6,0.1,0.1,0.1,0.1]$, and the uniform vector $[0.2,\dots,0.2]$. All six baselines are designed for single-scalar optimization, so we adapt each to optimize $\vw^\top\vv$ over all five objectives, with a new run per preference. \textbf{This protocol deliberately favors the baselines.} Each baseline targets the exact weights it is scored at, whereas \method{} never sees them during search and serves all eleven from one tree.

\paragraph{Models and prompts.}
We test \method{} and all six baselines on GPT-5 Mini \citep{singh2026openaigpt5card}, Claude Haiku 4.5 \citep{anthropic2025haiku45}, Gemini 3 Flash Preview \citep{gemini3flash2025}, and DeepSeek-V4-Flash \citep{deepseekai2026deepseekv4highlyefficientmilliontoken}. Within a run, a single model makes every LLM call: it proposes at most $b=4$ actions on each visit to a decision node, realizes the selected action into at most $N_r=3$ role specializations, and executes each finished workflow at temperature 0.7.

\section{Experimental Results and Analysis}\label{sec:experiments}
\begin{table}[t]
  \centering
  \small
  \setlength{\tabcolsep}{3pt}
  \caption{\textbf{Held-out hypervolume of the front served across eleven testing preferences.} All methods use GPT-5 Mini. Each baseline$^{\dagger}$ is adapted to the five objectives and rerun once per preference, targeting the exact weight it is later scored at. \method{} instead serves all eleven preferences from a single search that never saw them. \textbf{Even so, \method{} attains the best hypervolume on four of six benchmarks and the best mean.} \textbf{Bold} marks the best per column, \underline{underline} the second best. Mean is reported with standard deviation across the six benchmarks.}
  \label{table:main}
  \begin{tabular}{lcccccc c}
  \toprule
  Method & AIME 2026 & MATH-Hard & MBPP & SWE-Bench & GPQA-D & HotpotQA-H & Mean \\
  \midrule
  AFlow$^{\dagger}$      & 29.85 & \textbf{21.87} & 27.71 & 7.11 & 17.41 & 18.57 & 20.42\pmstd{8.17} \\
  MaAS$^{\dagger}$       & 24.61 & 14.28 & 21.37 & 7.00 & 15.13 & 24.69 & 17.85\pmstd{6.97} \\
  FlowSteer$^{\dagger}$  & 18.47 & 17.34 & 19.86 & 4.06 & 12.58 & 9.68  & 13.66\pmstd{6.07} \\
  SkillFlow$^{\dagger}$  & 18.55 & 15.75 & 10.77 & 0.81 & 10.35 & 4.89  & 10.19\pmstd{6.59} \\
  ADAS$^{\dagger}$       & 27.71 & 19.84 & 26.50 & 6.84 & \underline{20.17} & 28.89 & 21.66\pmstd{8.22} \\
  EvoMAS$^{\dagger}$     & 28.68 & 20.74 & 22.59 & \underline{7.30} & 20.14 & \textbf{30.35} & 21.63\pmstd{8.19} \\
  \midrule
  \method{} (Online)  & \underline{31.39} & 21.69 & \underline{34.92} & \textbf{7.54} & 18.42 & \underline{30.14} & \underline{24.02}\pmstd{10.19} \\
  \method{} (Offline) & \textbf{33.02} & \underline{21.85} & \textbf{35.14} & 6.89 & \textbf{20.39} & 29.27 & \textbf{24.43}\pmstd{10.42} \\
  \bottomrule
  \end{tabular}
\end{table}
  
\paragraph{Example.}
\autoref{fig:example} shows one \method{} search on SWE-Bench building a front of workflows. Each trial samples a fresh preference $\vw$ and rolls out one workflow, so the hypervolume of the non-dominated set rises whenever a new trade-off enters the front. The front ranges from a single \texttt{PatchRepair} at 25k tokens to a four-step workflow that branches, votes, and iteratively repairs at 194k. Every construction trace records how each action changed the five objectives, so practitioners can see which actions played the most significant role in shaping the final performance.

\paragraph{Main results.} \autoref{table:main} reports the hypervolume of the front each method serves across the eleven testing preferences. \method{} is strongest overall, with its two variants achieving the two best mean hypervolumes.\footnote{Hypervolume magnitude depends on the reference point $\vr_0$ but the {\em relative} ranking across methods provides an unbiased comparison.}
We attribute this advantage to cross-preference transfer. A baseline improves its solution only for the preference it optimizes. \method{} also searches under one sampled preference per trial, but CHVI backs up each outcome as a vector available to all preferences, so every preference benefits from the entire search.

\paragraph{Online and offline \method.}
The two \method{} variants in \autoref{table:main} reach nearly the same hypervolume at very different search costs. Online \method{} executes every generated workflow, so execution accounts for over 99\% of search tokens: a single search spends 132.3M execution tokens on AIME 2026 and 134.9M on MBPP (\autoref{table:cost}), while the baselines average 5.6--28.7M per preference. Offline \method{} instead pretrains $\fth$ on the execution data those online searches already collected across the six tasks, so its search runs no workflow and costs only 2.9--3.6M proposer and realizer tokens, about 2\% of the online cost and a significant saving compared with the baselines. Yet the offline version attains slightly better hypervolume, showing that once enough execution data is available, \method{} can search effectively with a pretrained evaluator.

\paragraph{\method{} outperforms on nearly every objective, not just in hypervolume.}
Hypervolume summarizes all five objectives into one score, so \autoref{fig:by-dataset} breaks the comparison out by objective. Each axis shows the best value a method attains across the eleven testing preferences. \method{} matches or beats every preference-tuned baseline on nearly every objective and benchmark.

\subsection{Ablation Study}
\label{ssec:ablation}
\begin{table}[t]
  \centering
  \small
  \setlength{\tabcolsep}{3.5pt}
  \renewcommand{\arraystretch}{0.93}
  \caption{\textbf{Held-out hypervolume under four base models, where \method{} is the only method that ranks in the top two on all four.} Each method is rerun end to end with the tested model.}
  \label{table:models}
  \begin{tabular}{lcccc cc}
  \toprule
  Method & GPT-5 Mini & Haiku 4.5 & Gemini 3 Flash & DeepSeek-V4-Flash & Mean & Worst \\
  \midrule
  AFlow$^{\dagger}$      & \underline{29.85} & 14.54 & \textbf{30.33} & 19.86 & 23.64\pmstd{7.75} & 14.54 \\
  MaAS$^{\dagger}$       & 24.61 & 14.40 & 17.88 & 17.77 & 18.66\pmstd{4.28} & 14.40 \\
  FlowSteer$^{\dagger}$  & 18.47 & 15.78 & 28.75 & 13.61 & 19.15\pmstd{6.70} & 13.61 \\
  SkillFlow$^{\dagger}$  & 18.55 & \textbf{16.80} & 2.39  & 14.68 & 13.11\pmstd{7.32} & 2.39  \\
  ADAS$^{\dagger}$       & 27.71 & 16.02 & 29.19 & \textbf{22.90} & \underline{23.95}\pmstd{5.93} & \underline{16.02} \\
  EvoMAS$^{\dagger}$     & 28.68 & 7.60  & 10.83 & 15.33 & 15.61\pmstd{9.27} & 7.60  \\
  \midrule
  \method{} (Online) & \textbf{31.39} & \underline{16.09} & \underline{30.15} & \underline{22.13} & \textbf{24.94}\pmstd{7.19} & \textbf{16.09} \\
  \bottomrule
  \end{tabular}
\end{table}
\paragraph{\method{} is generalizable across models.}
\autoref{table:models} reruns the AIME 2026 comparison under three additional base models. \method{} ranks first on GPT-5 Mini and second on the other three and the best overall (the mean). No other method stays in the top two on all four. Each baseline that leads one column falls behind on another: SkillFlow leads on Claude Haiku 4.5 but collapses to 2.39 on Gemini 3 Flash Preview, AFlow leads on Gemini but falls 2.27 below \method{} on DeepSeek-V4-Flash, and ADAS leads on DeepSeek but trails elsewhere. In practice, deployments fix the base model for reasons unrelated to workflow quality, so the worst-case score matters most, and \method{} ranks first there (16.09).

\begin{wraptable}{r}{0.5\textwidth}
  \vspace{-\baselineskip}
  \centering
  \small
  \setlength{\tabcolsep}{3.5pt}
  \caption{\textbf{Action-selection ablation on AIME 2026 with Claude Haiku 4.5.} CZT attains the best in-sample hypervolume at the lowest cost.}
  \label{table:czt}
  \begin{tabular}{lcc}
  \toprule
  Action selection & In-sample HV $\uparrow$ & Tokens (M) $\downarrow$ \\
  \midrule
  Chebyshev       & 17.06 & 43.76 \\
  Hypervolume-UCB & 17.87 & 61.01 \\
  Pareto-UCB      & 16.80 & 26.44 \\
  CZT (ours)      & \textbf{20.35} & \textbf{23.99} \\
  \bottomrule
  \end{tabular}
  \vspace{-\baselineskip}
\end{wraptable}
\paragraph{Preference-conditioned selection matters.} We replace CZT at the decision node with the three other action-selection rules of \S\ref{sec:related_tree_search_morl}, leaving every other component of \method{} unchanged. Hypervolume-UCB and Pareto-UCB each reduce an action's value set to a single scalar that ignores the sampled preference. Chebyshev scalarization does condition on $\vw$, but keeps one global estimate per action across the entire simplex. All three lose 12--17\% of the in-sample hypervolume (on the build set under $\vr_0$) while spending more tokens. CZT keeps separate estimates per simplex region and refines only where the winning action changes, yielding a better front at lower cost.

\paragraph{Query-level search does not generate a better workflow than task-level search.} \method{} also supports query-level search, which builds a small tree per query instead of one tree per task. We run both modes on all six benchmarks under the same setting. Building the twenty query-level trees costs 0.6--0.9$\times$ the execution tokens of one task-level tree, yet the workflows they produce consume 1.1--11.6$\times$ the tokens and 1.1--6.6$\times$ the calls per problem, with no gain on other objectives. The reason is that query-level search is more likely to overfit, with a held-out to in-sample hypervolume ratio as low as 0.28, compared with at least 0.66 for task-level trees (\autoref{table:cost}). Query-level search is thus unnecessary. Our finding echoes the recent finding of \citet{wang-etal-2026-always}, who show that a small set of top-ranked task-level workflows already covers as many queries as a per-query generator.

\section{Conclusion}\label{sec:conclusion}
We presented \method{}, a multi-objective workflow generator that formulates workflow generation as an MOMDP and solves it with Convex-Hull MCTS. Its core idea is to value every node by a set of reachable trade-offs rather than a scalar, so a single search approximates the Pareto front and serves any preference by lookup. Across six benchmarks and four base models, \method{} consistently ranks in the top two in average hypervolume. We envision \method{} as a foundation for agentic systems that replace one-size-fits-all pipelines with a Pareto landscape of trade-offs, searched once and queried at inference time for each user's choice of quality, cost, and trust. Appendix~\ref{app:limitations} discusses limitations.


\bibliography{ref}
\bibliographystyle{iclr2027_conference}

\newpage
\appendix

\etocdepthtag.toc{appendix}

\begingroup
\etocsettagdepth{mainmatter}{none}
\etocsettagdepth{appendix}{subsection}
\etocsetstyle{section}{}{}
  {\vskip3pt plus1pt\noindent\hangindent1.9em\hangafter1
   {\bfseries\makebox[1.9em][l]{\etocnumber}\etocname}\nobreak
   \leaders\hbox to 6pt{\hfil.\hfil}\hfill\nobreak{\bfseries\etocpage}\par}
  {\vskip2pt}
\etocsetstyle{subsection}{}{}
  {\noindent\hspace*{1.9em}\hangindent4.7em\hangafter1
   \makebox[2.8em][l]{\etocnumber}\etocname\nobreak
   \leaders\hbox to 6pt{\hfil.\hfil}\hfill\nobreak\etocpage\par}
  {}
\etocsettocstyle{\section*{Appendix}\vskip4pt}{}
\parindent0pt \parskip0pt
\tableofcontents
\endgroup

\section{Additional Results}
\label{app:additional}

\subsection{Accuracy-Cost Pareto Fronts}
\label{app:fronts}

\begin{figure}[ht]
\centering
\includegraphics[width=\textwidth]{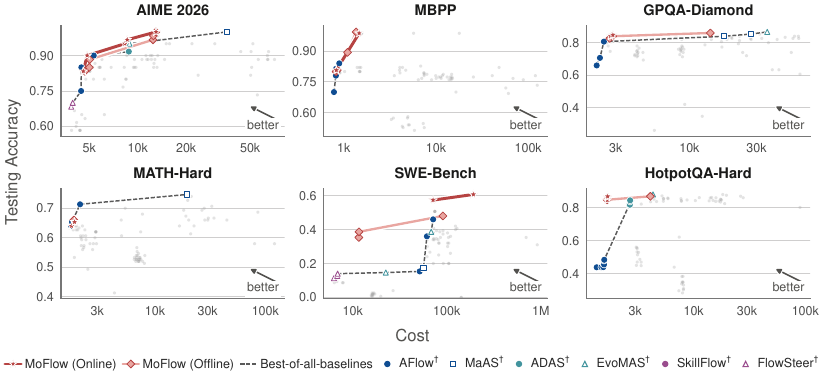}
\caption{\textbf{Accuracy-cost Pareto fronts on the six benchmarks.} Test accuracy against tokens per problem under GPT-5 Mini (up and left is better). The dashed front combines all workflows generated by the six adapted baselines$^{\dagger}$, each rerun once per preference. Gray points are dominated workflows. \textbf{\method{}, from a single search that never saw the eleven preferences, matches or extends this combined baseline front on five of the six benchmarks}, trailing only on MATH-Hard, where the baselines' most accurate workflow stays ahead at roughly ten times the cost.}
\label{fig:fronts}
\end{figure}

\autoref{fig:fronts} plots each method's Pareto front on accuracy versus cost, the pair among our five objectives that practitioners care about most in deployment.

\subsection{Task-Level versus Query-Level Search}
\label{ssec:modes-exp}

\begin{figure}[ht]
\centering
\includegraphics[width=\textwidth]{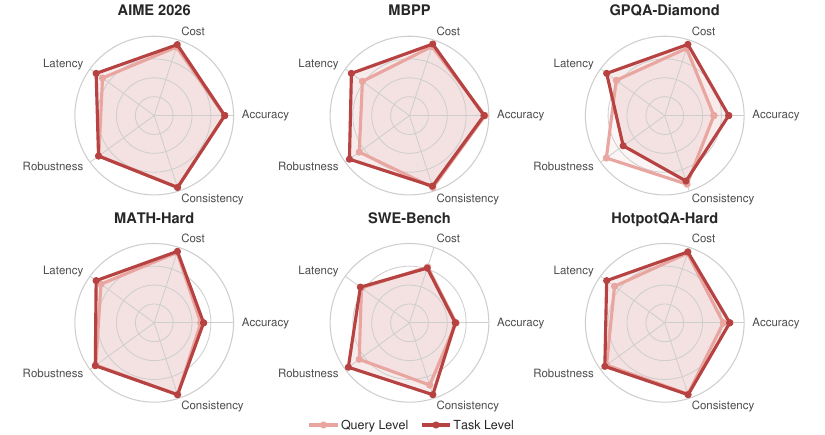}
\caption{\textbf{Per-objective performance of task-level versus query-level search.} One tree per benchmark (\emph{task-level}) against one small tree per query (\emph{query-level}), under the same setting as \S\ref{ssec:setup}. Each axis shows the best value a search mode attains across its eleven served workflows, with cost and latency inverted so that outward is always better. \textbf{Task-level search matches or exceeds query-level search on nearly all objectives across the six benchmarks.}}
\label{fig:modes}
\end{figure}

\begin{table}[ht]
\centering
\footnotesize
\setlength{\tabcolsep}{3.2pt}
\caption{\textbf{Cost comparison between task-level and query-level search.} Search cost is the execution tokens spent building each tree. Inference cost and latency are the mean tokens and sequential calls per problem, averaged over the workflows returned for the eleven preferences. $\rho_{\mathrm{HV}}$ is the hypervolume generalization ratio (Eq.~\ref{eq:hvratio}). Query-level trees are cheaper to build, but the workflows they return spend up to 11.6$\times$ the tokens and 6.6$\times$ the calls per problem at inference. \textbf{Query-level search overfits the inputs it was built on: $\rho_{\mathrm{HV}}$ falls as low as 0.28 on unseen draws of the same query, against at least 0.66 for every task-level tree.}}
\label{table:cost}
\begin{tabular}{l rr rr rr rr}
\toprule
& \multicolumn{2}{c}{Search cost (M tokens)} & \multicolumn{2}{c}{Inference cost (tokens)} & \multicolumn{2}{c}{Inference latency (calls)} & \multicolumn{2}{c}{$\rho_{\mathrm{HV}}$ ($\uparrow$)} \\
\cmidrule(lr){2-3}\cmidrule(lr){4-5}\cmidrule(lr){6-7}\cmidrule(lr){8-9}
Benchmark & Task & Query & Task & Query & Task & Query & Task & Query \\
\midrule
AIME 2026     & 132.3 & 111.9 & 5.6k   & 23.0k  & 1.36 & 4.27 & 0.68 & 0.37 \\
MATH-Hard     & 115.7 & 74.1  & 5.7k   & 12.5k  & 1.00 & 3.80 & 0.93 & 0.73 \\
MBPP          & 134.9 & 85.4  & 2.1k   & 20.1k  & 1.19 & 7.89 & 0.87 & 0.28 \\
SWE-Bench     & 597.2 & 452.1 & 102.7k & 115.3k & 6.53 & 7.36 & 0.66 & 0.63 \\
GPQA-Diamond  & 232.0 & 162.7 & 2.8k   & 32.5k  & 1.00 & 6.25 & 1.95 & 0.93 \\
HotpotQA-Hard & 122.8 & 92.2  & 5.6k   & 15.8k  & 1.36 & 5.59 & 0.86 & 0.65 \\
\bottomrule
\end{tabular}
\end{table}

\method{} runs at one of two granularities. \emph{Task-level} search builds one tree per benchmark on its build set and serves every testing preference from that tree by lookup. \emph{Query-level} search builds a small front for each incoming query, scoring candidates against the query's paraphrases and repeated runs. Appendix~\ref{app:modes} derives the complexity of both modes. \autoref{fig:modes} compares the best value each mode reaches on every objective, and \autoref{table:cost} reports build cost, per-problem inference cost, and generalization. We measure generalization by the \emph{hypervolume generalization ratio}
\begin{equation}
\rho_{\mathrm{HV}}=\frac{\mathrm{HV}_{\text{held-out}}}{\mathrm{HV}_{\text{in-sample}}},
\label{eq:hvratio}
\end{equation}
where $\mathrm{HV}_{\text{in-sample}}$ is a tree's hypervolume on the inputs its search evaluated and $\mathrm{HV}_{\text{held-out}}$ its hypervolume on unseen inputs. For a task-level tree these are the build and held-out sets of \S\ref{ssec:setup}. For a query-level tree, the in-sample inputs are the paraphrases and repeated runs scored during search, the held-out inputs are unseen paraphrases, and \autoref{table:cost} reports the ratio of mean hypervolumes across a benchmark's query trees. $\rho_{\mathrm{HV}}$ near one means the front generalizes, and a value well below one means the search overfit. Task-level GPQA-Diamond ($\rho_{\mathrm{HV}}=1.95$) exceeds one because its 20-question build split is harder than the 50-question held-out split (the same generated workflows score 0.63 accuracy in-sample but 0.83 held out), and hypervolume compounds this per-objective gap multiplicatively.

\subsection{Full Search Tree Visualizations}\label{app:trees}
\begin{figure}[ht]
\centering
\includegraphics[width=\textwidth]{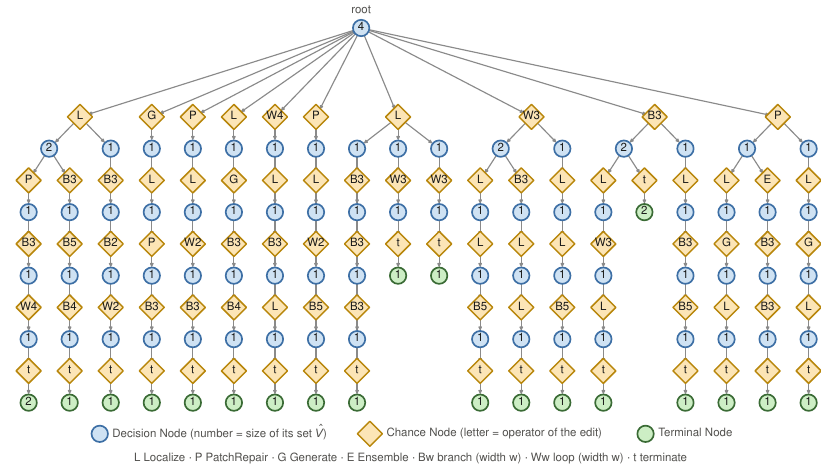}
\caption{\textbf{A complete query-level search tree on SWE-Bench,} built with GPT-5 Mini. The tree contains 90 decision nodes (blue), 83 chance nodes (gold), and 20 evaluated terminals (green). The number in each circle is the size of the node's CCS $\Vhat(s)$: a 1 means the node holds a single trade-off. Two terminals show a 2 because one of their three repeated runs solved the problem while another failed but used fewer tokens or calls, so neither run dominates. Each terminal stores these pruned per-run returns in $\Vhat(s)$, while Eq.~\ref{eq:chance-backup} backs up their average $\E[R]$. The tree thus records that a workflow can fail and fail cheaply.}
\label{fig:tree-swe}
\end{figure}

\begin{figure}[ht]
\centering
\includegraphics[width=\textwidth]{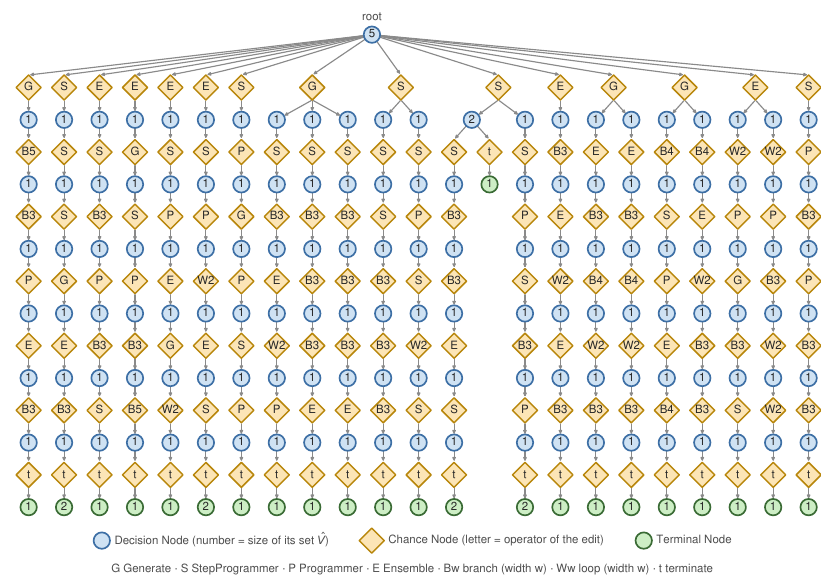}
\caption{\textbf{A complete query-level search tree on MATH-Hard.} It contains 156 decision nodes, 148 chance nodes, and 23 evaluated terminals. The root keeps five trade-offs, ranging from an accurate workflow at 10.8k tokens down to a single-call workflow at 1.0k tokens that fails, and the blue ``2'' at depth 1 shows that an interior node can also keep more than one trade-off. The search favors code execution, selecting \texttt{StepProgrammer} or \texttt{Programmer} in 45 of its 148 actions.}
\label{fig:tree-math}
\end{figure}

\begin{figure}[ht]
\centering
\includegraphics[width=\textwidth]{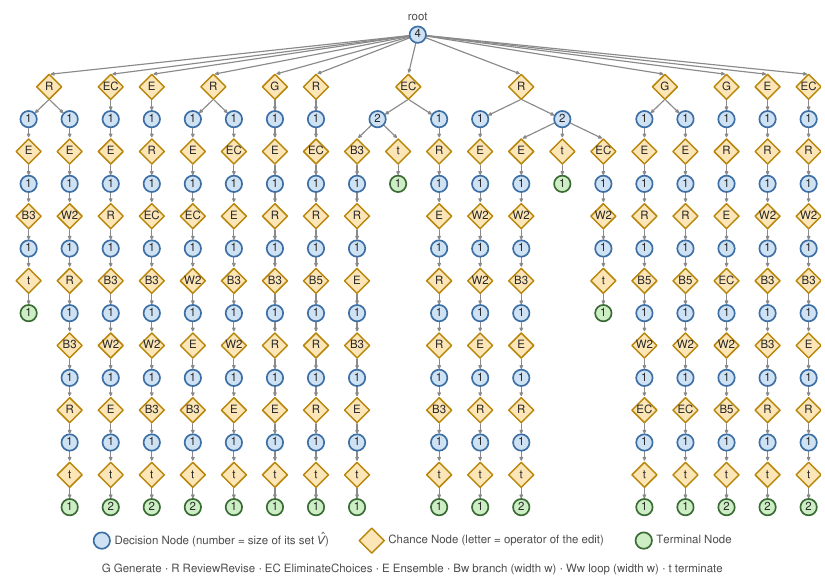}
\caption{\textbf{A complete query-level search tree on GPQA-Diamond.} The tree contains 122 decision nodes, 116 chance nodes, and 20 evaluated terminals. The root spans the widest cost range of the three trees: its accurate workflow costs 82k tokens and 14 calls, while its two-call workflows stay under 9k tokens. Seven of the 20 terminals show a 2, each keeping both its cheapest accurate run and its cheapest failed run. The tree thus records that the same workflow can succeed, or fail at a slightly lower cost. The search is critique-driven, selecting the \texttt{ReviewRevise} operator in 42 of its 116 actions.}
\label{fig:tree-gpqa}
\end{figure}

Figures~\ref{fig:tree-swe}--\ref{fig:tree-gpqa} show three query-level trees in full, one per task type: code repair (SWE-Bench), mathematics (MATH-Hard), and scientific QA (GPQA-Diamond). We visualize query-level rather than task-level trees because the latter, at roughly 240--490 decision nodes, are too large to render completely.

\clearpage

\section{Limitations}
\label{app:limitations}
We note three limitations of \method{}. First, it compares value vectors through the linear score $\vw^\top\vv$, so it covers the convex hull of the Pareto front but not its concave regions, which only nonlinear scalarizations (e.g., Chebyshev) can reach. Second, like AFlow \citep{zhang2025aflow}, it requires a manually designed operator pool for each task. With a large open pool, the trial budget alone could not substitute for a stronger proposer. Third, to save cost, every online search stops early once the hypervolume of the executed front plateaus (Appendix~\ref{app:setup-details}), a rule that halted all reported runs between trials 40 and 50, so we never observe \method{} (online) under a substantially larger budget. Moreover, the fixed operator pool and horizon bound the structurally distinct workflows the search can reach (even though free-form roles leave the raw workflow set unbounded), so additional trials would likely revisit the front rather than extend it.

\section{\method{} Details}\label{app:method}

Appendix~\ref{app:momdp} gives the state and action spaces of the MOMDP together with the prompt templates, and Appendix~\ref{app:algo} the full search procedure.
\subsection{Construction MOMDP and Prompt Templates}\label{app:momdp}\label{app:prompts}

\paragraph{State representation.}
A state is an immutable \texttt{WorkflowGraph}. Two partial workflows with the same structure share a canonical key and map to the same node, so a workflow reachable by different edit sequences is visited only once.

\paragraph{Action space.}
An action is one of three atomic edits.
\begin{itemize}[leftmargin=*]
\item \texttt{AddOperator($o$, width, role)} appends operator $o$ from the task's operator pool (below). The \texttt{role} is a free-text label that names the node's prompt archetype (e.g., ``step-by-step solver'' or ``code verifier'').
\item \texttt{AddControl(branch$|$loop, width)} wraps an operator in a control structure: \texttt{branch} runs \texttt{width} parallel attempts and keeps the best, while \texttt{loop} refines sequentially for \texttt{width} rounds.
\item \texttt{Terminate} closes the workflow.
\end{itemize}
Roles are not edited independently. The proposer assigns an initial role when it proposes each node-adding edit, and the realizer rewrites that role into up to $N_r$ distinct specializations when the edit is instantiated (realizer prompt below). An edit is \emph{legal} only if it yields a valid workflow: a terminal workflow cannot be extended, \texttt{Terminate} requires a non-empty graph whose last operator emits an answer (e.g., \texttt{Generate} or \texttt{Ensemble}), widths are bounded, and depth cannot exceed the horizon $H$. These constraints define the legal action set $A(s)$.

\paragraph{Operator pools.}
Below we show the operators we defined for each benchmark.
\begin{itemize}[leftmargin=*]
\item \textbf{AIME 2026, MATH-Hard}: \texttt{Generate}, \texttt{Ensemble}, \texttt{Programmer}, \texttt{StepProgrammer};
\item \textbf{MBPP}: \texttt{Generate}, \texttt{Ensemble}, \texttt{ReviewRevise}, \texttt{TestCode};
\item \textbf{SWE-Bench}: \texttt{Generate}, \texttt{Localize}, \texttt{PatchRepair}, \texttt{Ensemble}, \texttt{Custom};
\item \textbf{GPQA-Diamond}: \texttt{Generate}, \texttt{Ensemble}, \texttt{ReviewRevise}, \texttt{EliminateChoices};
\item \textbf{HotpotQA-Hard}: \texttt{Generate}, \texttt{Ensemble}, \texttt{Custom}, \texttt{Decompose}, \texttt{EvidenceSelect}, \texttt{GroundCheck}.
\end{itemize}
We purposefully design operators aligned with those of existing baselines to ensure a fair comparison.

\paragraph{Prompt templates.}
Three templates drive the proposer, the realizer, and the executor, and all three are fixed across the four models and six benchmarks. The \textbf{proposer} template returns candidate actions at a decision node.
\begin{promptbox}{Proposer Prompt}
\tt \small
You are designing an LLM workflow (a DAG of operators) to solve the following task:
\\
\{task\}
\\ \\
The current partial workflow has these operators in order: \{ops\}.
\\ \\
Available operators (choose only from these, by exact name):
\\
\{catalogue\}
\\ \\
Propose up to \{n\} good NEXT atomic edits to extend or finish the workflow. An edit is one of: a base operator; "loop" (wrap an operator to refine it sequentially for `width` passes); "branch" (run `width` parallel attempts of an operator and keep the best); or "Terminate".
\\ \\
Pick operators and roles that fit THIS task; the last operator before Terminate must produce the task's required answer/output (do not finish on an analysis-only step).
\\ \\
Reply with ONLY a JSON list, each item \{"op": <operator|"loop"|"branch"|"Terminate">, "operator": <inner operator for loop/branch, else omit>, "width": <int >= 1>, "role": <short label>, "why": <one-line rationale>\}.
\end{promptbox}

The \textbf{realizer} template turns one selected action into $N_r$ distinct successors.
\begin{promptbox}{Realizer Prompt}
\tt \small
You are instantiating one step of an LLM workflow for the following task:
\\
\{task\}
\\ \\
The workflow so far has these operators in order: \{ops\}.
\\ \\
The next step adds a `\{operator\}` operator (width \{width\}). That operator's job: \{operator\_desc\}
\\ \\
Propose \{n\} DISTINCT concrete ways to play that step -- each a different role/approach the operator could take FOR THIS TASK (a distinct prompt archetype, reasoning strategy, or specialization), so that executing them could plausibly lead to different outcomes. Each role must fit the operator's job and this task.
\\ \\
Reply with ONLY a JSON list of \{n\} items, each \{"role": <short distinct approach label, <= 40 chars>, "why": <one-line rationale>\}.
\end{promptbox}

The \textbf{executor} assembles each operator call from four parts in order: the benchmark's \emph{task description}, the problem behind an \texttt{Input:} header, the \emph{operator-specific instruction}, and the benchmark's \emph{answer format}. Only the task description and answer format depend on the benchmarks, while the operator instruction is fixed and the realizer fills in its \{role\} slot. The example below shows the assembled prompt for a \texttt{Generate} step on MATH-Hard.
\begin{promptbox}{Executor Prompt (\texttt{Generate} on MATH-Hard)}
\tt \small
You are an expert competition mathematician. Solve the problem carefully, reasoning step by step.
\\ \\
Input:
\\
\{problem\}
\\ \\
Role: \{role\}. Reason step by step, then give your best answer.
\\ \\
Put your final answer in \textbackslash boxed\{...\} on the last line. Give it in exact, simplest form (reduced fraction, exact radical, or integer), not a decimal approximation.
\end{promptbox}

\subsection{\method{} Search Procedure}
\label{app:algo}

\paragraph{Set operations.}
The backups of Eqs.~\ref{eq:chance-backup} and~\ref{eq:decision-backup} use three set operations. \emph{Prune} discards every vector dominated by another in the set. \emph{Union} merges two sets and then prunes. The Minkowski sum $X\msum Z=\prune\{x+z:x\in X,z\in Z\}$ adds an immediate reward set to a future value set. Because prune removes only Pareto-dominated vectors, each node stores the full Pareto front, a superset of the minimal CCS.

\paragraph{Optimistic chance backup.}
The union over successors in Eq.~\ref{eq:chance-backup} is deliberately optimistic rather than exact. The exact CHVI backup \citep{10.1145/1390156.1390162,ICAPS20paper100} is the probability-weighted Minkowski sum over successors. It values an action by averaging over the realizations of the edit. Our union instead keeps every vector achievable under at least one realization. When a preference $\vw$ takes the scalarized maximum of this set, the action is credited with its best realization, as though the search could choose the outcome. In reality, the search only observes which realization occurs, so the union upper-bounds the exact value for every $\vw$. We accept the optimistic bias for two reasons. The union is much cheaper, since the exact sum multiplies front sizes across successors while the union only merges them (Appendix~\ref{app:modes}). More importantly, the bias affects only which workflow is generated, while every reported number is an executed return.

\paragraph{CZT metric.}
CZT covers the context--arm space $\gP_s=\Delta^{D-1}\times A(s)$ with balls under the metric $d_s((\vw,a),(\vw',a'))=\|\vw-\vw'\|_\infty$ if $a=a'$ and $U$ otherwise, where $V_{\max}$ is the largest possible scalarized return and $U\ge V_{\max}$ keeps distinct actions incomparable \citep{ICAPS20paper100}. Each arm is identified by its edit's property tuple (operator, width, and role for \texttt{AddOperator}, or control kind, width, and wrapped operator for \texttt{AddControl}), so a re-proposed edit joins an existing ball only when the tuples match.

\paragraph{Selection and zooming.}
On each visit with context $\vw$, CZT scores the relevant balls by the index of Eq.~\ref{eq:czt-index}. A ball is relevant if its action was proposed on this visit and $\vw$ lies in the ball's domain, the portion of $B$ not covered by a finer same-action ball. Because distinct actions sit at distance $U\ge V_{\max}$, only same-action balls can tighten the min in Eq.~\ref{eq:czt-index}, so a well-explored neighbor constrains the optimism of an under-explored ball. The visit applies the action of the largest-index ball and folds the observed scalarized reward into $\nu(B)$. Once $\mathrm{conf}(B)\le r(B)$, where $\mathrm{conf}(B)=\kappa\sqrt{\log N/(1+n(B))}$ for a ball selected $n(B)$ times under trial budget $N$, the ball activates a half-radius child of the same action centered at $\vw$.

\section{Implementation Details}
\label{app:setup-details}

\autoref{table:hyperparams} lists all \method{} hyperparameters and reproducibility details. The rest of this section describes the motivation behind certain design and hyperparameter choices.

\paragraph{Dataset and evaluation.}
We split each benchmark into a build set and a held-out set. Each problem carries $m=3$ paraphrases, generated by GPT-5 Mini at temperature 0.9 with a prompt that varies only wording so the gold answer is unchanged. Grading uses the boxed answer on the two math suites, reference tests on MBPP, the official harness on SWE-Bench, the gold letter on GPQA-Diamond, and token-level F1 on HotpotQA-Hard. All methods share this grading, and a single trial measures all five objectives: consistency comes from the variance of accuracy across the $J$ repeated runs, and robustness from its variance across the $m$ paraphrases. The factor 4 in $\mathrm{clip}(1-4\Var(\mathrm{acc}),0,1)$ is not tuned. Per-problem accuracy lies in $[0,1]$, so its variance is at most $1/4$ and the factor simply rescales the score to $[0,1]$. Every LLM call is routed through the LiteLLM proxy server \citep{litellm_software}, and cost sums the per-call token usage the proxy meters. Latency counts sequential LLM calls rather than wall-clock time, which varies with provider, hardware, and load, so parallel calls within an \texttt{Ensemble} or \texttt{branch} count as one round.

\paragraph{Reference point.}
Hypervolume is meaningful only when every served workflow dominates the reference point $\vr_0$, because a workflow that falls outside it contributes zero volume and can distort the method ranking. The most expensive workflow served in our experiments uses $1.9{\times}10^5$ tokens per problem and the slowest makes 12.5 sequential calls, both on SWE-Bench. We set the cost and latency coordinates of $\vr_0$ to $3{\times}10^5$ tokens and 20 calls, leaving ample headroom beyond these observed extremes.

\paragraph{Baseline adaptation.}\label{app:baseline-adapt}
We adapt each baseline by changing only the score it optimizes, replacing its scalar objective with $\vw^\top\vv$ and rerunning the method once per testing preference. All other aspects of the baseline (design, implementation, prompts, and hyperparameters) remain untouched. Every method searches on the same build set, and every workflow is scored under the shared protocol ($J=3$ repeated runs and $m=3$ paraphrases), so each objective is measured identically across methods.

\paragraph{Uncertainty-gated objective evaluation.}
Accuracy, robustness, and consistency can only be measured from a finished workflow. At a partial workflow, $\fth$ instead predicts these three objectives along with its own uncertainty. We execute only the nodes where that uncertainty exceeds the gate, which in practice triggers at the leaves of a small fraction $\gamma$ of trials. At a gated node, we let a completion policy $\pi_c$ fill the open slots of $s$ until termination, execute a diverse beam of $k$ completions, and replace the prediction with their non-dominated vectors,
\begin{equation}
\Vhat_{\mathrm{exec}}(s)\approx
\prune\big(\{R(\mathrm{exec}(c)):c\in\mathrm{Beam}_k(\pi_c,s)\}\big).
\label{eq:beam}
\end{equation}
The executed returns also feed the replay buffer on which $\fth$ is trained, so even when $\fth$ is confidently wrong at some node and the gate never fires there, the finished workflows that extend it are still executed and retraining on their results corrects the prediction.

\paragraph{Total cost.}\label{app:total-cost}
Each baseline rerun receives the method's default budget (5.6--28.7M execution tokens per preference, depending on method and benchmark), so a baseline spends roughly 62--316M tokens per benchmark across the eleven preferences. A single online \method{} search uses 116--597M tokens, the same order as one baseline sweep, while offline \method{} serves all eleven preferences from just 2.9--3.6M proposer and realizer tokens. Across the full study, the \method{} searches behind \autoref{table:main}, \autoref{table:czt}, \autoref{table:models}, and \autoref{table:cost} consume 3.1B metered tokens over about 450{,}000 execution calls. Including the estimated 9B tokens for baseline reruns, the study totals roughly 12B tokens, about \$14{,}700 at September 2026 on-demand rates.

\begin{table}[t]
  \centering
  \small
  \setlength{\tabcolsep}{5pt}
  \caption{\textbf{\method{} hyperparameters and reproducibility information.} All settings are shared across the six benchmarks and four base models except the horizon and set sizes, noted in their rows.}
  \label{table:hyperparams}
  \begin{tabular}{ll}
  \toprule
  Name & Value \\
  \midrule
  \multicolumn{2}{@{}l}{\emph{Construction MOMDP and search}} \\
  Number of objectives $D$ & 5 \\
  Horizon $H$ & 6 (4 on SWE-Bench) \\
  Trial cap $N$ (online) & 100; from trial 40, stop once 10-trial HV gain $<\!1\%$ \\
  Trial budget $N$ (offline) & 300, no early stop \\
  Action proposals per visit $b$ & 4 \\
  Realizations per action $N_r$ & 3 \\
  Completion-beam width $k$ & 3 \\
  CZT confidence constant $\kappa$ & 4 \\
  Random seed & 0 \\
  \midrule
  \multicolumn{2}{@{}l}{\emph{Objectives and evaluation}} \\
  Repeated runs $J$ & 3 per problem \\
  Paraphrases $m$ & 3 \\
  Execution temperature & 0.7 \\
  Reference point $\vr_0$ & $[0,\,-3{\times}10^5,\,-20,\,0,\,0]$ \\
  Testing preferences & 11 fixed $\vw$ \\
  Build\,/\,held-out set sizes & $10/20$ on AIME 2026, $20/50$ elsewhere \\
  \midrule
  \multicolumn{2}{@{}l}{\emph{Value model $\fth$}} \\
  Architecture & 2-layer message-passing GNN, width 64, role dim 8 \\
  Role encoder & all-MiniLM-L6-v2 \\
  Predicted objectives & accuracy, robustness, consistency \\
  Uncertainty gate & 0.15 \\
  Online refit & every 32 replay entries; Adam, lr $10^{-2}$, 60 epochs \\
  Offline training & Adam, lr $10^{-3}$, decay $10^{-4}$, 600 epochs, patience 100 \\
  Offline training pool & 801 executed workflows from six online searches \\
  Offline train\,/\,val split & $70\%/30\%$ by workflow hash ($558/243$ workflows) \\
  \midrule
  \multicolumn{2}{@{}l}{\emph{Compute}} \\
  Execution tokens (online) & 116--597M per search \\
  LLM tokens (offline) & 2.9--3.6M per search \\
  Total study cost & ${\sim}$12B tokens, ${\sim}$\$14{,}700 \\
  \bottomrule
  \end{tabular}
\end{table}

\clearpage
\section{\method{} Search Complexity Analysis}\label{app:modes}\label{ssec:modes}

\method{} searches at two granularities, defined and compared in Appendix~\ref{ssec:modes-exp}. Task-level search pays for one tree per task and serves every later preference from it, while query-level search pays for a new small front on each incoming query. This section derives the cost of a single trial and extends it to both granularities.

\paragraph{Cost of one trial.}\label{ssec:trialcost}
The \emph{executed front} is the non-dominated set of return vectors from all workflows executed so far in the search. Let $N$ be the number of trials, $H$ the horizon, $b$ the per-visit proposal budget, $N_r$ the realizations per selected action, $D$ the number of objectives, $k$ the completion-beam width, $F$ the maximum per-node front size after pruning, $|\pi|$ the number of tree nodes in the policy unrolled for one preference, $\Phi$ the maximum number of vectors on the executed front during the search, $\beta$ the maximum number of active CZT balls at a decision node, $E$ the cost of one workflow evaluation, and $\ell\ll E$ the cost of one proposer or realizer LLM call. Because the proposer redraws at most $b$ actions on every visit, a decision node's branching factor is not $b$ but the number $\bar A\le\min\{|A(s)|,\,bT_s\}$ of distinct actions ever registered there, where $T_s$ is the number of visits to node $s$. The rollout, workflow execution, set backups, executed-front update, and CZT selection sum to
\begin{equation}
\underbrace{O\!\big((1+\gamma k)E+H\ell\big)}_{\text{executions and LLM calls}}
+\underbrace{O\!\big(H(\bar A+N_r)^2F^2D\big)}_{\text{set backups}}
+\underbrace{O\!\big((1+\gamma k)\Phi D\big)}_{\text{executed-front update}}
+\underbrace{O\!\big(H\beta^2D\big)}_{\text{CZT selection}},
\label{eq:trialcost}
\end{equation}
with $\gamma$ the fraction of trials whose rollout ends at an uncertain interior node and so triggers a completion beam. The beam's completion policy adds $O(kH)$ proposer calls of its own on those trials, absorbed into $O(H\ell)$ since $k$ is a constant. The first term dominates, since executing workflows costs far more than pruning small vector sets, except in the offline mode, which executes no workflows while building the tree and incurs only the $O(H\ell)$ proposer and realizer calls.

\paragraph{Derivation of Eq.~\ref{eq:trialcost}.}
The four terms follow one trial in the order its costs are paid. The rollout visits at most $H$ decision nodes and makes about two LLM calls at each. One proposer call redraws candidate actions and one realizer call samples the chosen action's outcome, so the rollout pays $O(H\ell)$. The execution cost then depends on how the trial ends. A terminal leaf is executed once at cost $E$, and a confident interior leaf is valued by $\fth$ with no execution. An uncertain interior leaf triggers a completion beam, which executes its $k$ completions at cost $kE$. A trial therefore executes workflows at most $1+\gamma k$ times. Combined with the rollout's LLM calls, this yields the first term, $O\big((1+\gamma k)E+H\ell\big)$. After execution, the set backups run once at every level of the visited path, and their cost comes from pruning. Discarding Pareto-dominated vectors from $n$ candidates in $\R^D$ takes $O(n^2D)$ (one $O(D)$ dominance check per pair) \citep{Roijers_2013}. The chance backup of Eq.~\ref{eq:chance-backup} unions the fronts of at most $N_r$ realized successors, each of size at most $F$, so it prunes at most $N_rF$ candidates at cost $O(N_r^2F^2D)$. The decision backup of Eq.~\ref{eq:decision-backup} likewise unions the $\bar A$ registered actions' fronts, so it prunes at most $\bar AF$ candidates at cost $O(\bar A^2F^2D)$. Since $\bar A^2+N_r^2\le(\bar A+N_r)^2\le2(\bar A^2+N_r^2)$, summing the two backups over the at most $H$ levels gives the second term, $O\big(H(\bar A+N_r)^2F^2D\big)$. The trial's $O(1+\gamma k)$ executed return vectors then enter the executed front. Inserting one vector runs one $O(D)$ dominance check against each of the at most $\Phi$ stored vectors, so the executed-front update gives the third term, $O\big((1+\gamma k)\Phi D\big)$. Finally, choosing an action at each visited node evaluates the active CZT balls using the index of Eq.~\ref{eq:czt-index}. Each index takes a minimum over the at most $\beta$ active balls, and each pair of balls requires one $O(D)$ distance evaluation. One selection therefore costs $O(\beta^2D)$, and the at most $H$ selections give the fourth term, $O\big(H\beta^2D\big)$.

\paragraph{Task-level search.}
We run $N$ trials once and store the resulting executed front and tree, at $N$ times Eq.~\ref{eq:trialcost},
\[
O\!\big(N(1+\gamma k)E+NH\ell\big)
+O\!\big(NH(\bar A+N_r)^2F^2D\big)
+O\!\big(N(1+\gamma k)\Phi D\big)
+O\!\big(NH\beta^2D\big).
\]
When actions have several realizations, inference unrolls the conditional policy by scanning, at each of its $|\pi|$ nodes, the $\bar A$ registered actions' fronts of size at most $F$, at $O(|\pi|\bar AFD)$. When realizations are unique, inference is a single scan of the executed front, at $O(\Phi D)$.

\paragraph{Query-level search.}
The system receives a query $q$ and runs a smaller search for it. A single query gives only a binary success signal, so we evaluate workflows on a small neighborhood of paraphrases and repeated samples. Accuracy is then the mean success rate on the original query, robustness is stability across paraphrases, and consistency is stability across repeated runs. The cost is Eq.~\ref{eq:trialcost} with $N_q$ in place of $N$,
\[
O\!\big(N_q(1+\gamma_q k)E_q+N_qH\ell\big)
+O\!\big(N_qH(\bar A+N_r)^2F^2D\big)
+O\!\big(N_q(1+\gamma_q k)\Phi_qD\big)
+O\!\big(N_qH\beta^2D\big),
\]
where $E_q$ is one evaluation over the query neighborhood, $\Phi_q$ the maximum number of vectors on that query's executed front during its search, and $\gamma_q$ the gated fraction of the query's trials. Unlike in task-level search, the front itself is the answer. Across many queries the replay buffer for $\fth$ is shared, so later queries may need fewer completion beams and fewer trials.

\paragraph{Comparison with the baselines.}
Every baseline spends its entire budget on a single trade-off, implicit or fixed, so a new preference $\vw$ forces either a fresh search (AFlow, ADAS, EvoMAS) or a fresh training run (MaAS, FlowSteer, SkillFlow). Covering a set $\gW$ of testing preferences therefore multiplies each method's full budget by $|\gW|$, e.g.\ $O(|\gW|NE)$ for the search-based generators. \method{} instead pays $O\big(N((1{+}\gamma k)E+H\ell)\big)$ once in task-level search, and only its $O(NH\ell)$ proposer and realizer portion in the offline mode. Every later preference is then answered with zero LLM calls, either by unrolling the conditional policy at $O(|\pi|\bar AFD)$ or, when realizations are unique, by a single scan of the executed front at $O(\Phi D)$.

\end{document}